\documentclass[english]{IEEEtran}
\usepackage[T1]{fontenc}
\usepackage[latin9]{inputenc}
\usepackage{color}
\usepackage{babel}
\usepackage{array}
\usepackage{textcomp}
\usepackage{mathrsfs}
\usepackage{multirow}
\usepackage{amsmath}
\usepackage{amssymb}
\usepackage{graphicx}
\usepackage[unicode=true]
 {hyperref}

\makeatletter

\providecommand{\tabularnewline}{\\}

\newcommand{\lyxaddress}[1]{
	\par {\raggedright #1
	\vspace{1.4em}
	\noindent\par}
}

\@ifundefined{showcaptionsetup}{}{%
 \PassOptionsToPackage{caption=false}{subfig}}
\usepackage{subfig}
\makeatother

\begin{document}
\title{Attention-based Context Aggregation Network for Monocular Depth Estimation}
\author{Yuru Chen, Haitao Zhao, Zhengwei Hu}
\maketitle

\lyxaddress{School of Information Science and Engineering, East China University
of Science and Technology, China}
\begin{abstract}
Depth estimation is a traditional computer vision task, which plays
a crucial role in understanding 3D scene geometry. Recently, deep-convolutional-neural-networks
based methods have achieved promising results in the monocular depth
estimation field. Specifically, the framework that combines the multi-scale
features extracted by the dilated convolution based block (atrous
spatial pyramid pooling, ASPP) has gained the significant improvement
in the dense labeling task. However, the discretized and predefined
dilation rates cannot capture the continuous context information that
differs in diverse scenes and easily introduce the grid artifacts
in depth estimation. In this paper, we propose an attention-based
context aggregation network (ACAN) to tackle these difficulties. Based
on the self-attention model, ACAN adaptively learns the task-specific
similarities between pixels to model the context information. First,
we recast the monocular depth estimation as a dense labeling multi-class
classification problem. Then we propose a soft ordinal inference to
transform the predicted probabilities to continuous depth values,
which can reduce the discretization error (about 1\% decrease in RMSE).
Second, the proposed ACAN aggregates both the image-level and pixel-level
context information for depth estimation, where the former expresses
the statistical characteristic of the whole image and the latter extracts
the long-range spatial dependencies for each pixel. Third, for further
reducing the inconsistency between the RGB image and depth map, we
construct an attention loss to minimize their information entropy.
We evaluate on public monocular depth-estimation benchmark datasets
(including NYU Depth V2, KITTI). The experiments demonstrate the superiority
of our proposed ACAN and achieve the competitive results with the
state of the arts. The source code of ACAN can be found in \href{https://github.com/miraiaroha/ACAN}{https://github.com/miraiaroha/ACAN}
\end{abstract}

\section{Introduction}

Depth information has a significant impact on understanding 3D scenes
and can benefit the tasks such as 3D reconstruction\cite{Silberman2012Indoor}
, 3D object detection \cite{Simon2018Complex}, visual simultaneous
localization and mapping (SLAM) \cite{Tateno2017CNN,laina2016deeper},
and autonomous driving \cite{Geiger2012Are}. Estimating the pixel-wise
depth of scenes from RGB images has triggered wide research recently
in the computer vision community. The goal of depth estimation is
to assign each pixel in an image the distance between the observer
and the scene point represented by this pixel. Estimating the depth
from a single monocular image is ill-posed without any geometric cues
or priors. Therefore the previous works mainly focus on the stereo
vision \cite{Hirschm2005Accurate,Roberts2011Structure}, in which
the binocular images or multi-view images are adopted to obtain the
disparity map, and the depth information can be further reconstructed
from the disparity map by utilizing the camera parameters. However,
the drawbacks of stereo matching lie in the blind areas of the prediction
due to the existence of occlusion, and the predicted results might
be distorted by inaccurate camera parameters.

Recently, deep-neural-network-based methods have been widely used
in computer vision tasks and achieved great performances. Convolutional
neural networks (CNNs) have been proved effective for image classification.
Simultaneously, people have applied CNN to dense labeling tasks, such
as monocular depth estimation \cite{Eigen2014Depth,Eigen2014Predicting},
semantic segmentation \cite{Chen2018DeepLab,Zhao2016Pyramid} and
edge detection \cite{Xie2015Holistically} by modifying the network
structure of CNN.

Despite the above success, there still has existed some key challenges
in monocular depth estimation tasks. In common deep-CNN-based image-processing,
the spatial scales of feature maps continue to shrink as the network
goes deeper due to the successive pooling and stride operations, which
allows the deep CNN to learn the increasingly abstract representations
and fuse the global features to obtain the image-level prediction.
However, this translation invariance property may hinder the dense
prediction tasks, such as semantic segmentation and depth estimation,
where detailed spatial information and image structure are crucial.
To overcome this problem, some previous works utilize the skip connection\cite{Ronneberger2015U}
to combine the feature maps produced by shallow layers and deep layers
of the same spatial scales. Moreover, the intermediate supervision
\cite{Newell2016Stacked,Wei2016Convolutional} is applied to the multi-scale
cues to progressively refine the prediction. In other works \cite{Yu2017Dilated,yu2015multi},
the application of dilated convolution maintains the resolution while
extending the receptive fields and without introducing extra parameters.

Another challenge comes from the depth distribution of objects in
the scene. Huang et al. \cite{Huang2000Statistics} studied the statistics
of range images of natural scenes (called depth maps in the depth
estimation field), which showed that the range images can be decomposed
into piecewise smooth regions that show little dependencies with each
other and the sharp discontinuities typically exist in the object
boundaries. Therefore, the concept of \textquotedblleft objects\textquotedblright{}
in the scene can be better defined in terms of changes in depth rather
than some low-level features, such as color, intensity, texture, lighting
etc. From this perspective, depth estimation as a classification task
can be regarded as a generalized semantic segmentation task while
the labels between pixels are not independent. Accordingly, the key
point in depth estimation is how to capture the long-range context
information of intra-object and inter-object. Yu et al. \cite{yu2015multi}
used serialized layers with increasing dilation rates to extend the
receptive fields of convolutional kernels, while the research works
\cite{Chen2018DeepLab,Chen2017Rethinking} implement an \textquotedblleft atrous
spatial pyramid pooling (ASPP)\textquotedblright{} framework to capture
multi-scale objects and context information by placing multiple dilated
convolution layers in parallel. However, the discretized and limited
dilation rates cannot cover the complex and continuous scene depth
and easily introduce the grid artifacts \cite{Wang2018Understanding},
which can be found in Fig. \ref{fig9}.

\begin{figure*}
\begin{centering}
\includegraphics[scale=0.35]{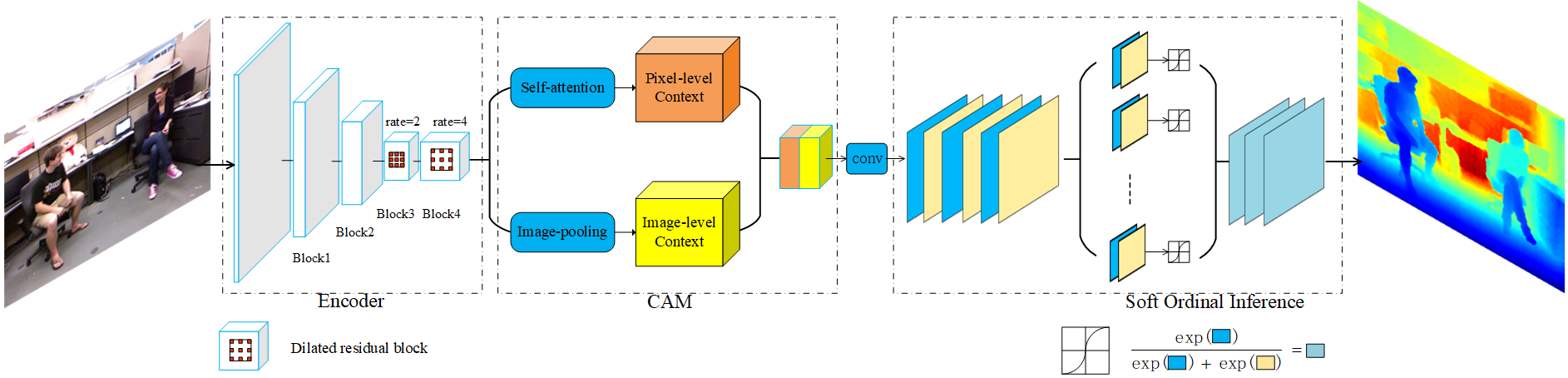}
\par\end{centering}
\caption{Network architecture. The ResNet is adopted by the ACAN as the encoder,
where the cascaded 2-dilated and 4-dilated convolutions are used to
avoid the over-downsampling. In the decoder, the CAM is proposed to
extract and aggregate both the pixel-level and image-level context.
Finally, our proposed soft ordinal inference will translate the predicted
probabilities into continuous depth values.\label{fig1}}
\end{figure*}
In view of the challenges, this paper proposes a novel depth estimation
algorithm, called the attention-based context aggregation network
(ACAN), to tackle the monocular depth estimation problem. The deep
residual architecture \cite{he2016deep} is adopted by ACAN, where
dilated convolutions are used to maintain the spatial scale. To extract
the continuous pixel-level context information, the self-attention
module \cite{Vaswani2017Attention,Wang2017Non,liu2015parsenet} is
plugged into our model to approximate depth distribution of scenes
by learning an attention map that carries the normalized similarities
between all the pixels. According to the learned attention map, we
can obtain the context information of each pixel. Different from prefixed
or prestructured local kernels, our proposed attention model can obtain
adaptive similarities, which reflect the relationships between each
pixel and any other pixels in the whole feature map. Instead of using
predefined regions and extracting sparse context information in ASPP,
the proposed ACAN can learn the attention weights associated with
meaningful contextual areas, resulting in predicting the piecewise
smooth depth. The comparison between ASPP and our proposed ACAN can
be seen in Fig. \ref{fig2}. To reduce the inconsistency between RGB
image and depth map, KL divergence is adopted to model the divergence
between the distribution produced by the self-attention model and
the distribution constructed by the corresponding ground truth depth.
To further incorporate the image-level information for depth estimation,
the image-pooling \cite{Chen2017Rethinking,liu2015parsenet} is utilized
in this paper. Finally, our proposed soft ordinal inference translates
the predicted probabilities into the continuous depth values and produce
more realistic transitional regions.

The main contributions of this paper can be summarized as follows:
\begin{itemize}
\item We propose a pixel-level attention model for the monocular depth estimation
that can capture the context information associated with each pixel.
In addition, the aggregation of pixel-level context and image-level
context is effective to promote the estimation performance. Our experimental
results demonstrate that the proposed pixel-level attention model
outperform the ASPP based model since the generated pixel-level context
information of ACAN is flexible and continuous, and therefore avoid
the grid effect.
\item To eliminate the large semantic gap from 2D image texture and depth
map, we introduce KL divergence as our attention loss to minimize
the divergence between the distribution of the attention map and the
distribution of the similarity map constructed by the ground truth
depth. The effectiveness of the attention loss is confirmed by our
ablation experiments.
\item An easy-implemented soft inference strategy is proposed in this paper,
which can reduce the discretization error and produce more realistic
depth map compared with the naïve hard inference.
\end{itemize}

\section{Related Work}

Estimating the depth of a scene is a traditional task in computer
vision and has been studied for a long period. As a pioneering work,
Saxena et al. \cite{Saxena2005Learning} infer the depth from monocular
cues based on Markov Random Field (MRF), and further develop their
method in \cite{Saxena2007Learning}, where the smoothness assumption
is imposed to the superpixels to enforce the neighboring constraint.
Their work later extended for the 3D model generation \cite{Saxena20083}.
In \cite{Liu2010Single}, semantic labels are incorporated into the
MRF framework to guide the depth estimation. Ladicky et al. \cite{Ladicky2014Pulling}
showed that the property of perspective geometry could be used to
learn a much simpler classifier to predict the likelihood of a pixel
instead of a pixel-wise depth classifier. All these works provide
novel thoughts, while most of them rely on strong geometric constraints
and hand-crafted features thus limit their models to generalize to
diverse scenarios.

Recently, a large body of works adopts the deep neural network for
monocular depth estimation\cite{Eigen2014Depth,Eigen2014Predicting,Hu2018Revisiting,Chen2017Rethinking,laina2016deeper,yan2018monocular}.
The seminal work of Eigen et al. \cite{Eigen2014Depth} first proposed
a multi-scale coarse-to-fine model, where the fine network refines
the global prediction from coarse network to produce a more detailed
result, and the innovative scale-invariant loss is proved an effective
loss function both for training and evaluation. They then extended
their model to a three-scale architecture for three dense labeling
tasks, i.e. predicting normal, label and depth \cite{Eigen2014Predicting}.
In order to solve the heavy-tailed effect of depth values reported
in \cite{Roy2016Monocular}, Laina et at. \cite{laina2016deeper}
presented that the reverse Huber loss \cite{Zwald2012The} is more
appropriate than standard L2 regression loss for depth estimation
since Huber loss is more sensitive to small errors. While the deep
CNN-based methods are excellent at extracting image features, they
are weak in reconstructing high-resolution images due to the down-sampling
operation and lack of structural constraints, therefore, often obtain
the depth estimation with distorted boundaries and counterfeit regions.
To tackle this problem, Hu et al. \cite{Hu2018Revisiting} proposed
the notable loss function whose three items are complementary with
each other and the loss function is edge-aware. Garg et al \cite{Garg2016Unsupervised}
proposed an unsupervised framework for single view depth estimation
with a photometric reconstruction loss between stereo pairs. Under
this setting, Godard et al. \cite{Godard2017Unsupervised} further
proposed a combination of an L1 loss and the structural similarity
index (SSIM) term \cite{Zhou2004Image} as the reconstruction loss
and explicitly imposed a spatial smoothness constraint \cite{Heise2014PM}
for the synthesized image. Chen et al. \cite{chen2018rethinking}
regarded the depth estimation as an image-to-image translation task,
additionally utilized an adversarial loss with the discriminator as
a structural penalty.

Besides the above methods using the task-specific loss or geometric
prior to supervise the network learning, there exists another research
route that fuses multi-scale information in CNNs for pixel-level prediction
\cite{Xie2015Holistically,Ronneberger2015U,Yu2017Dilated,yu2015multi,Chen2017Rethinking,Kim2018Deep}.
Most of them applied an encoder-decoder architecture, where a reliable
encoder adaptively learns the hierarchical features of input RGB images.
In the decoder, the specially designed building blocks are employed
to recover the spatial resolution or leverage the multi-scale context
to restore the finer details. Laina \cite{laina2016deeper} introduced
an up-sampling block to improve the output resolution. In the research
works \cite{Xu2017Multi,Liu2015Deep,Li2015Depth,Liu2015Learning,yan2018monocular},
the conditional random field (CRF) based models have been utilized
for the multi-scale features to estimate the fine-grained depth maps.
Kim et al. \cite{Kim2018Deep} proposed a deep variational model that
integrates the predictions from the global and local networks. In
the research works \cite{Ronneberger2015U,Godard2017Unsupervised,Zhang2018Progressive,Li2018Monocular},
skip connections were added to concatenate the detail-abundant features
from the encoder with the decoder features of the corresponding scales.
Although these works give the impressively sharp inferences, they
also introduce the inevitable artifacts in some highly textured regions
\cite{Zhang2018Progressive,moukari2018deep}. To address this problem,
Fu et al. \cite{fu2018deep} employed the dilated convolutions to
capture context information in multiple scales, a typical example
is ASPP \cite{Chen2017Rethinking}, which has been well studied in
semantic segmentation \cite{Chen2018DeepLab,yu2015multi,Chen2017Rethinking}.
While the dilated-convolution-based methods have achieved the state-of-the-art,
the dilated kernels introduce a sparse sub-sampling of activations,
which results in an inherent problem identified as \textquotedblleft gridding\textquotedblright{}
\cite{Chen2017Rethinking}.

To deal with the gridding problem, different from using prefixed structures
of the dilated kernels, we design an attention model to extract the
continuous multi-scale context by adaptively learning the pixel-level
similarity map. The output features of the decoder can be computed
by a weighted sum of contextual regions, which is essential for the
fine-grained depth estimation. Moreover, due to our designed attention
loss, the ambiguity caused by the large semantic gap could be partly
eliminated and the produced attention map could be task-specific.
CRF is widely adopted to obtain the pixel-level pairwise similarities
as the context information \cite{Chen2018DeepLab,Xu2017Multi,Zheng2015Conditional,Lin2015Efficient,Li2015Depth,Liu2015Learning}.
However, the similarities are patch-wise and only able to compute
between the pixel and its local neighborhoods. Armed with the pixel-level
attention, which could be regarded as a global structural extractor,
our proposed ACAN can capture the long-range dependencies of intra-object
by directly computing interactions between widely scattered pixels
which share similar depth values.

Although estimating a depth range is more robust than estimating a
depth value for each pixel and the classification strategy can put
different weights on different depth ranges according to the tasks
of depth estimation \cite{Cao2017Estimating}, the naïve hard-threshold-based
depth inference ignores the predicted confidence of the depth distribution
and usually introduces the stepped artifacts \cite{fu2018deep,Cao2017Estimating}.
In this paper, taking the full advantage of the output confidence
of the proposed network, we propose a soft inference strategy to reduce
the discretization error and eliminate the stepped artifacts.

\section{Methods}

This section introduces the architecture of our proposed ACAN and
associated loss functions for the monocular depth estimation, which
maps an RGB image to its corresponding depth map in an end-to-end
fashion. 

\subsection{Network Architecture}

The network architecture is illustrated in Fig. \ref{fig1}, which
also uses the encoder-decoder framework. We consider the ResNet as
the encoder to extract the dense features of RGB image. ResNet shows
great gradient-propagating capability in deeper networks by adding
identity branches to plain network, which is essential for depth estimation
due to its large receptive field \cite{laina2016deeper}. However,
the over-downsampling of original ResNet may hinder the reconstruction
of the fine-grained depth map. Instead, we replace the block3 and
block4 in ResNet with 2-dilated and 4-dilated residual blocks, which
favor for the initialization of pre-trained parameters and maintain
the scale of the subsequent feature maps \cite{Chen2017Rethinking,yu2015multi}.

In the decoder, we propose a novel building block that called the
context aggregation module (CAM) to enable the network to capture
the discriminative image-level and pixel-level context information.
We finally jointly train our model using the combination of attention
loss and ordinal loss. We then describe CAM and the training losses
in detail.

\subsection{Context Aggregation Module }

\begin{figure}
\subfloat[self-attention module\label{fig2(a)}]{\begin{centering}
\hspace{1.2cm}\includegraphics[scale=0.3]{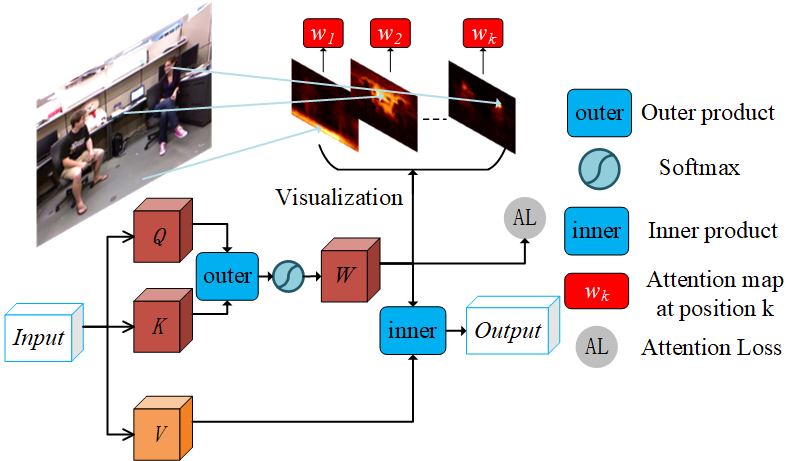}
\par\end{centering}
}

\subfloat[ASPP\label{fig2(b)}]{\begin{centering}
\hspace{1.5cm}\includegraphics[scale=0.4]{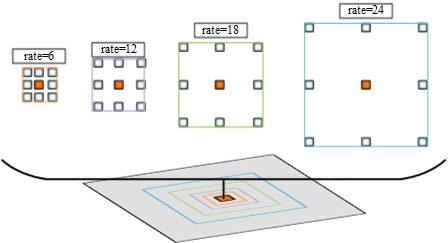}
\par\end{centering}

}

\caption{(a) Our pixel-level attention model can capture the global and dense
context for each location, while ASPP only parallelizes a limited
amount of convolution kernels thus resulting in the sparse sampling.
In addition, the attention loss is proposed to reduce the semantic
gap between RGB image and depth map.\label{fig2}}

\end{figure}
As illustrated in Fig. \ref{fig1}, the CAM includes two branches.
The top branch is a pixel-level attention model, i.e. self-attention,
the bottom branch is the image pooling operation. In the end, the
resulting output features from the two branches are concatenated and
passed to the subsequent classifier.

\textbf{Self-Attention}: The self-attention module maps a query and
a set of key-value pairs to an output, where query, key and value
represent three feature vectors extracted by the input via three transformation
functions respectively. The output is computed as a weighted sum of
the values in the feature space, where the weight assigned to each
value is computed by a pairwise function of the query with the corresponding
key. The details of the self-attention model are illustrated in Fig.
\ref{fig2(a)}.

Specifically, as demonstrated in Fig. \ref{fig2(a)}, the feature
map $x\in\mathbb{R}^{N\times C_{in}}$ inputted to the self-attention
module is first encoded into two embedded features, i.e. key feature
$K\in\mathbb{R}^{N\times C_{K}}$ and query feature $Q\in\mathbb{R}^{N\times C_{Q}}$,
where $K=\phi(x)$, $Q=\varphi(x)$, $\phi$ and $\varphi$ are the
transformation functions, $N$ is the number of spatial positions,
i.e. $N=H\times W$, and $C_{K}=C_{Q}<C_{in}$. The normalized attention
weight $w_{i,j}$ can be computed by a pairwise function $\mathscr{\mathcal{F}}$
as follows,

\begin{equation}
w_{i,j}=\frac{1}{\Omega_{i}(x)}\mathscr{\mathcal{F}}(x_{i},x_{j})\ \ i,j=1,\ldots,N\label{eq:(1)}
\end{equation}

Where $\Omega_{i}(x)$ is the normalized factor, defined as $\Omega_{i}(x)=\Sigma_{j=1}^{N}\mathscr{\mathcal{F}}(x_{i},x_{j})$
. To be more specific, we consider the embedded Gaussian as the pairwise
function.

\begin{equation}
\mathscr{\mathcal{F}}\left(x_{i},x_{j}\right)=e^{\frac{\phi(x_{i})^{T}\varphi(x_{j})}{\sqrt{C_{K}}}}\label{eq:(2)}
\end{equation}

where $\sqrt{C_{K}}$ is the scaling factor to prevent values produced
by outer-product operation growing large in magnitude, thus pushing
the attention weights into saturation. Therefore, the pixel-level
attention map $W\in\mathbb{R}^{N\times N}$ can be denoted equivalently
as

\begin{equation}
W=\mathrm{softmax}\left(Q^{T}K\right)\label{eq:(3)}
\end{equation}

By virtue of the learned attention weights, the output of self-attention
module at position $i$ can be defined as

\begin{equation}
c_{i}=\sum_{j=1}^{N}w_{i,j}\psi\left(x_{j}\right)\label{eq:(4)}
\end{equation}

where $\psi$ transforms $x$ into value feature $V\in\mathbb{R}^{N\times C_{V}}$.
By this way, the process of feature extraction is enhanced via explicitly
aggregating the context representation of the $i^{th}$ pixel according
to the learned attention weights. In this paper, we choose the $1\times1$
convolutions followed by the batch normalization layer and ReLU activation
function as $\phi$ and $\varphi$ and they share the same parameters.

\textbf{Image Pooling}: The image pooling has been widely used to
produce a class-specific activation map \cite{Zhou2015Learning}.
We first apply global average pooling (GAP) over the whole image to
reduce the 3D input feature maps to a 1D context vector, i.e., output
one response for every input feature map. Then by replicating feature
vector to the size of the input feature map, we can achieve the image-level
context map, which carries the mixture of information belonging to
different categories (channels) and helps to clarify local confusions
\cite{liu2015parsenet}. Essentially, we discover that the GAP is
similar to the channel-wise attention mechanism, the difference is
that the latter applies a softmax to the context vector produced by
the GAP to get an output probability. The effectiveness of image pooling
lies in its class-awareness. Given the input image of a scene, the
GAP can obtain its statistic prior to the features of the whole image.
Our experiment at Section \ref{subsec:4.5.3} confirms our assumption.

\subsection{Training Loss}

Our overall training loss $\mathcal{L}$ includes two items

\begin{equation}
\mathscr{\mathcal{L}=}\alpha_{att}L_{att}+\alpha_{ord}L_{ord}\label{eq:(5)}
\end{equation}

Where $L_{att}$ is the attention loss and $L_{ord}$ is the ordinal
loss, $\alpha_{att}$ and $\alpha_{ord}$ are the coefficients.

\textbf{Attention Loss}: As illustrated in Fig. \ref{fig2(a)}, to
bridge the semantic gap between the RGB image and depth, we consider
the KL divergence as the attention loss for training the attention
model, which measures the distance between the attention weights produced
by the self-attention with respect to its ground truth,

\begin{equation}
L_{att}=\frac{1}{N}\sum_{i=1}^{N}\sum_{j=1}^{N}w_{i,j}^{*}\ln\left(\frac{w_{i,j}^{*}}{w_{i,j}}\right)\label{eq:(6)}
\end{equation}

Where $w_{i,j}$ is the attention weights produced by Eq. \ref{eq:(1)}
and $w_{i,j}^{*}$ can be computed by the ground truth depth values
of $i^{th}$ pixel and $j^{th}$ pixel as follows,

\begin{equation}
w_{i,j}^{*}=\frac{\exp\left(\ln d_{\max}-|\ln d_{i}^{*}-\ln d_{j}^{*}|\right)}{\sum_{j=1}^{*}\exp\left(\ln d_{\max}-|\ln d_{i}^{*}-\ln d_{j}^{*}|\right)}\label{eq:(7)}
\end{equation}

where $d_{\max}$ denotes the preset value that is slightly larger
than the maximum depth value in the dataset. It is noted that $w_{i,\cdot}$
(the $i_{th}$ row of $w$) is the normalized attention distribution
of the $i^{th}$ pixel. Without using $L_{att}$, our attention model
can also produce certain plausible attention map according to the
extracted features of the image. However, this is problematic on the
highly textured surface as the assumption of appearance-depth correlation
is violated in these regions.

\textbf{Ordinal Loss}: The depth estimation is regarded as a pixel-level
classification problem. Due to the severe imbalance of depth data,
the samples are distributed more frequently in the small depth value
intervals \cite{Li2018Monocular}. However, since the magnitude of
the error of the large depth sample is larger than that of the small
depth sample, the network may over-fit the former. Hence, we discretize
the ground truth depth value $d_{i}^{*}$ in logarithmic space into
$K$ sub-intervals equally,

\begin{equation}
l_{i}^{*}=\lfloor\frac{\ln d_{i}^{*}-\ln d_{\min}}{\ln d_{\max}-\ln d_{\min}}\times K\rfloor\label{eq:(8)}
\end{equation}

Where $l_{i}^{*}\in\{0,1,\cdots,K-1\}$ is the quantified label of
$i^{th}$ pixel, $d_{i}^{*}$ is the continuous depth value of $i^{th}$
pixel. The ordered discretization thresholds $t^{k}\in\{0,1,\cdots,t^{K-1}\}$
can be obtained as follows,

\begin{equation}
t^{k}=e^{\ln d_{\min}+\frac{\ln d_{\max}-\ln d_{\min}}{K-1}*k}\label{eq:(9)}
\end{equation}

The ordinal loss \cite{Niu_2016_CVPR,fu2018deep} is adopted in the
proposed ACAN to learn our network parameters rather than the straightforward
cross entropy loss, which transfers the multi-class classification
problem into a series of simpler binary classification problems, each
of which only decides whether the sample is larger than $t^{k}$.
The ordinal loss imposes large loss on predictions that are not consistent
with the sequential property of the depth labels.

Formally, assuming $Y\in\mathbb{R}^{N\times2K}$ denotes the output
(confidence map) of the network. We can compute the ordinal loss at
spatial position $i$,

\begin{align}
\theta(y_{i}) & =-\sum_{k=0}^{l_{i}^{*}-1}\ln\mathscr{\mathcal{P}}_{i}^{k}-\sum_{k=l_{i}^{*}}^{K-1}\left(1-\ln\mathscr{\mathcal{P}}_{i}^{k}\right),\label{eq:(10)}\\
 & \mathscr{\mathcal{P}}_{i}^{k}=P(l_{i}>k)=\frac{e^{y_{i,2k+1}}}{e^{y_{i,2k}}+e^{y_{i,2k+1}}}\nonumber 
\end{align}

Where $l_{i}$ is the estimated label and $\mathscr{\mathcal{P}}_{i}^{k}$
is the ordinal probability that $l^{i}$ is larger than $k$ at position
$i$. The image-wise ordinal loss is defined as the average of $\theta(y_{i})$
over all spatial positions,

\begin{equation}
L_{ord}=\frac{1}{N}\sum_{i=1}^{N}\theta(y_{i})\label{eq:(11)}
\end{equation}

\textit{Soft Ordinal Inference}: Classification instead of regression
for depth estimation has been well studied in previous works, which
can naturally obtain the confidence of the depth distribution \cite{fu2018deep,Li2018Monocular,Cao2017Estimating}.
The element in the confidence map of each class only pays attention
to the specific depth interval, which simplifies the network learning.
However, it introduces the discretization error, which is sensitive
to the number of depth intervals. In addition, the hard-threshold-based
inference strategies \cite{fu2018deep,Cao2017Estimating} ignored
the obtained probability distribution which can be an important cue
during evaluating and may result in the step effect in the depth map,
reported in our experiment \ref{subsec:4.5.3}. Instead, we generalize
the naïve hard inference to a soft version, called the soft ordinal
inference to solve the above problems. The soft ordinal inference
takes full advantage of the confidence of predictions and shows a
strong ability to classify the transitional regions of inter-object.

After obtaining the probabilities of $K$ binary classification for
each pixel, the predicted depth $d_{i}$ of hard inference can be
computed as,

\begin{align}
d_{i} & =\frac{t^{l_{i}}+t^{l_{i+1}}}{2}\label{eq:(12)}\\
 & l_{i}=\sum_{k=0}^{K-1}\eta(\mathscr{\mathcal{P}}_{i}^{k}\geq0.5)\nonumber 
\end{align}

where $\eta(\cdotp)$ is an indicator function such that $\eta(true)=1$
and $\eta(false)=0$. The rounding operation of hard inference ignores
the probability (or confidence) predicted by the network, which may
distort the predictions of transitional regions that difficult to
distinguish.

However, our soft ordinal inference can transfer the predicted probabilities
to continuous depth values as follows,

\begin{align}
d_{i} & =\frac{t^{l_{i}}+t^{l_{i+1}}}{2}*(1-\mathcal{D}_{i})+\frac{t^{l_{i+1}}+t^{l_{i+2}}}{2}*\mathcal{D}_{i}\label{eq:(13)}\\
 & l_{i}=\lfloor\mathcal{s}_{i}\rfloor,\mathcal{D}_{i}=\mathcal{s}_{i}-l_{i}\nonumber \\
 & \mathcal{s}_{i}=\sum_{k=0}^{K-1}\mathscr{\mathcal{P}}_{i}^{k}\nonumber 
\end{align}

where $\lfloor\cdot\rfloor$ means the floor operation. $\mathcal{D}_{i}$
is between 0 and 1, which represents the extent to which the predicted
category is close to $l_{i+1}$. Actually, $\mathcal{s}_{i}$ is the
area under the probability distribution curve, which will be discussed
in Section \ref{subsec:4.5.3}.

\section{Experiments}

In this section, we investigate the performance of the proposed ACAN
model on two publicly available monocular depth datasets, NYU v2 Depth
\cite{Silberman2012Indoor}, KITTI \cite{Geiger2012Are}.

\subsection{NYU v2 Depth}

The original NYU v2 Depth dataset \cite{Silberman2012Indoor}consists
of around 240k RGB-D images of 464 indoor scenes, captured by a Microsoft
Kinect camera as video sequences. Following the research works \cite{Eigen2014Depth,laina2016deeper},
we use the official train\textbackslash test split, where 249 scenes
for training and 215 for testing. For training, we sample approximately
12k unique images with a fixed sampling frequency from each training
sequence and then fill in the invalid pixels of the depth map using
the colorization method, which is available in the toolbox of NYU
v2 dataset. The original image resolution is $480\times640$, we first
downsample it to $288\times384$ using bilinear interpolation and
then randomly crop to $256\times352$ pixels, as inputs to the network.
It is noted that the output of ACAN is 1/8 of ground truth depth in
scale, we upsample the output to the desired spatial dimension bilinearly.
Following \cite{Eigen2014Depth}, we use the same online data augmentation
strategies to increase the diversity of samples, which include random
scaling, random rotation, color, flips, and contrast. For testing,
we use the official 654 images and report our scores on a predefined
center cropping by Eigen \cite{Eigen2014Depth}.

\subsection{KITTI}

KITTI dataset \cite{Geiger2012Are} is composed of several outdoor
scenes captured by LIDAR sensor and car-mounted cameras while driving.
Following \cite{Eigen2014Depth}, we use the part of raw data selected
from the \textquotedblleft city\textquotedblright , \textquotedblleft residual\textquotedblright{}
and \textquotedblleft road\textquotedblright{} categories for training,
which including around 22k images from 28 scenes, and we evaluate
on 697 images selected from the other 28 scenes. The original resolution
is 375\texttimes 1242, and are resized to 160\texttimes 512 to form
the inputs. As the target depth maps projected by the point cloud
are sparse, we mask them out and evaluate the loss only on valid points
in both the training and testing phases.

\subsection{Implementation Details}

We implement our proposed model using the public deep learning framework
Pytorch on a single Nvidia GTX1080Ti GPU. In the proposed ACAN, both
ResNet-50 and ResNet-101 are the candidates for the encoder, whose
parameters are pretrained on the ImageNet classification task \cite{Russakovsky2015ImageNet}.
The depth intervals are set to 80 in all of our experiment. The learning
rate strategy applies a polynomial decay, which starts with the learning
rate of 2e-4 and is decayed with the power of 0.9 in the encoder.
Since the shallow convolution kernels are optimized well to extract
the general low-level features, we set the learning rate of the newly
added decoder layers to 10 times to that of the encoder layers. SGD
Optimization Algorithm is used to update the parameters, where momentum
and weight decay are set to 0.9 and 5e-4 respectively. We set the
weights of the different loss items to $\alpha_{att}$ and $\alpha_{ord}$=0.1.
The number of epoches is set to 50 both for KITTI and NYU v2, and
batch size is set to 8. We find that further increasing the iteration
number can hardly improve the performance.

\subsection{Evaluation Metrics}

Following previous works \cite{Eigen2014Depth}, we evaluate our depth
predictions using the following quantitative metrics:

Threshold: \% of $d_{i}$ s.t. $max\left(\frac{d_{i}}{d_{i}^{*}},\frac{d_{i}^{*}}{d_{i}}\right)=\delta<thr,thr=1.25,1.25^{2},1.25^{3}$

RMSE(linear): $\sqrt{\frac{1}{N}\sum_{i}||d_{i}-d_{i}^{*}||^{2}}$

RMSE(log): $\sqrt{\frac{1}{N}\sum_{i}||\ln d_{i}-\ln d_{i}^{*}||^{2}}$

Abs Relative Difference: $\frac{1}{N}\sum_{i}\frac{d_{i}-d_{i}^{*}}{d_{i}^{*}}$

Squared Relative Difference: $\frac{1}{N}\frac{|d_{i}-d_{i}^{*}|^{2}}{d_{i}^{*}}$

Please note that $N$ denotes the number of valid pixels.

\subsection{Discussion on Our Work}

In this subsection, we dig into the proposed context aggregation module
and the training loss for the proposed ACAN.

\subsubsection{Effect of the attention model and attention loss\label{subsec:4.5.1}}

\begin{figure}
\subfloat[example on NYU v2\label{fig3(a)}]{\begin{centering}
\hspace{1cm}\includegraphics[scale=0.6]{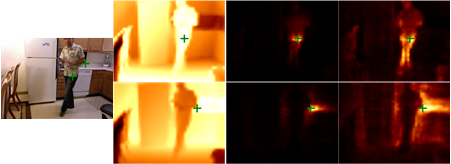}
\par\end{centering}

}

\subfloat[example on KITTI\label{fig3(b)}]{\begin{centering}
\includegraphics[scale=0.5]{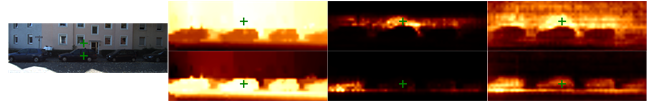}
\par\end{centering}

}

\caption{(a) example on NYU v2; (b) example on KITTI. The first column shows
the RGB images from the validation set. The second column presents
the ground truth contextual region computed by equation (7), the third
column and fourth column present the attention map produced by our
ACAN trained without and with L\_att respectively. The first and second
rows demonstrate the different attention maps located at \textquotedblleft \textit{\textcolor{green}{+}}\textquotedblright{}
of the same image.\label{fig3}}
\end{figure}

We first study the effectiveness of the proposed pixel-level attention
model. For qualitative analysis, we visualize the attention maps produced
by the attention model with and without attention loss respectively.
Fig. \ref{fig3} shows that (1) the pixel-level attention model does
predict the meaningful contextual regions which capture the long-range
dependencies, and is well adjusted to different scenarios adaptively.
(2) The visual comparison of the produced attention maps reveals that
the model trained with our $L_{att}$ can give the more detailed and
global contextual regions, which also acts as a structural extractor.
For example, in the first row of Fig. \ref{fig3(a)}, the attention
map without $L_{att}$ can only capture a local contextual region
at this location, while the attention map with $L_{att}$ highlights
the area of the standing man. Moreover, it also captures the context
of the stair that is away from the man but similar in depth, which
proves that the attention model can extract the contextual region
according to the task-specific semantical correlation rather than
the similarity of local and low-level features, i.e. the intensity
or texture. 

Quantitative results can be seen in Table \ref{table1}, where \textquotedblleft w/o\textquotedblright{}
denotes the model trained without the attention loss and \textquotedblleft w\textquotedblright{}
denotes the model trained with the attention loss. We can find that
our ACAN with attention loss can obtain a better performance in all
of the metrics.

\begin{table}
\centering{}%
\begin{tabular}{c|c|ccc|cc}
\hline 
\multicolumn{2}{c|}{} & $\delta_{1}$ & $\delta_{2}$ & $\delta_{3}$ & RMSE & ARE\tabularnewline
\hline 
\hline 
\multirow{2}{*}{NYU} & w/o  & 81.9\% & 95.8\% & 98.5\% & 0.502 & 0.140\tabularnewline
 & w  & 82.6\% & 96.4\% & 99.0\% & 0.496 & 0.138\tabularnewline
\hline 
\multirow{2}{*}{KITTI} & w/o  & 91.0\% & 98.0\% & 99.4\% & 3.902 & 0.090\tabularnewline
 & w  & 91.9\% & 98.2\% & 99.5\% & 3.599 & 0.083\tabularnewline
\hline 
\multicolumn{1}{c}{} & \multicolumn{1}{c}{} & \multicolumn{3}{c|}{higher is better} & \multicolumn{2}{c}{lower is better}\tabularnewline
\hline 
\end{tabular}\caption{Comparisons of ACAN trained with and without attention loss\label{table1}}
\end{table}

\subsubsection{Effect of the image-level feature}

We conduct the ablation experiment to reveal the effectiveness of
incorporating the image-level context to the proposed module. Results
are shown in Table \ref{table2}. All of these models are built on
ResNet-101. In the Table \mbox{III}, \textquotedblleft w\textquotedblright{}
represents an ACAN model with image-pooling block, \textquotedblleft w/o\textquotedblright{}
represents an ACAN model that sets the responses from GAP to a zero
vector for a comparison.

\begin{figure}
\subfloat[ Input RGB images from KITTI\label{fig4(a)}]{\begin{centering}
\includegraphics[scale=0.25]{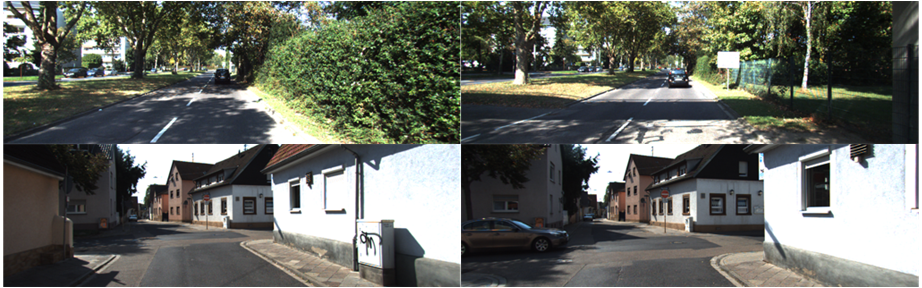}
\par\end{centering}

}\subfloat[ Matrix of L2 norm\label{fig4(b)}]{\begin{centering}
\includegraphics[scale=0.25]{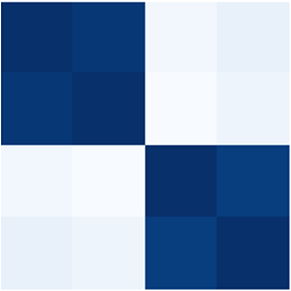}
\par\end{centering}

}

\caption{(a) input RGB images from KITTI, the images of the first row are from
\textquoteleft City\textquoteright{} category, and the images of the
second row are from \textquoteleft Residential\textquoteright{} category;
(b) L2 norm of image-level context vector of the four images.\label{fig4}}

\end{figure}

\begin{table}
\begin{centering}
\begin{tabular}{c|c|ccc|cc}
\hline 
\multicolumn{2}{c|}{} & $\delta_{1}$ & $\delta_{2}$ & $\delta_{3}$ & RMSE & ARE\tabularnewline
\hline 
\hline 
\multirow{2}{*}{NYU} & w/o  & 81.8\% & 96.1\% & 99.0\% & 0.504 & 0.140\tabularnewline
 & w  & 82.6\% & 96.4\% & 99.0\% & 0.496 & 0.138\tabularnewline
\hline 
\multirow{2}{*}{KITTI} & w/o  & 91.4\% & 98.2\% & 99.5\% & 3.733 & 0.085\tabularnewline
 & w  & 91.9\% & 98.2\% & 99.5\% & 3.599 & 0.083\tabularnewline
\hline 
\multicolumn{1}{c}{} & \multicolumn{1}{c}{} & \multicolumn{3}{c|}{higher is better} & \multicolumn{2}{c}{lower is better}\tabularnewline
\hline 
\end{tabular}
\par\end{centering}
\caption{Comparisons with and without image-level features on NYU and KITTI\label{table2}}
\end{table}
We further explore the effect of the image-pooling module by visualizing
the L2 norm matrix where each entry is calculated from the context
vectors produced by the GAP between the two images. As shown in Fig.
\ref{fig4}, the images are selected from KITTI dataset, each row
in Fig.\ref{fig4} is from the same scenario and is visually similar.
We observe that the image-pooling module has the remarkable distinguishability,
as it shows significant differences between different scenes and little
but not completely undifferentiated difference from the same scene.
It reveals that the image-pooling module can extract the discriminative
pattern of scenes. 

The above experiment reveals that the image-level context information
does act as a variant of channel-wise attention mechanism, which considers
the class-specific statistics prior that expresses the visual characteristic
of a scene. Therefore, our proposed ACAN is robust to the varied depth
samples from the dataset.

\subsubsection{Effect of the ordinal loss and soft ordinal inference\label{subsec:4.5.3}}

\begin{figure}
\subfloat[The plot of confusion matrix \protect \\
(ordinal inference)\label{fig5(a)}]{\begin{centering}
\includegraphics[scale=0.2]{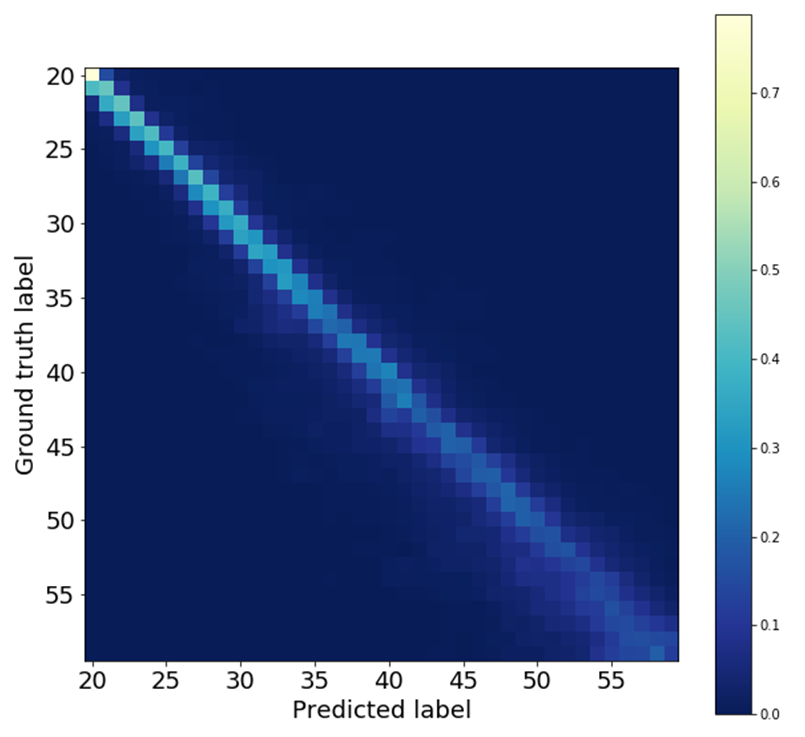}
\par\end{centering}
}\subfloat[The plot of confusion matrix \protect \\
(cross entropy)\label{fig5(b)}]{\begin{centering}
\includegraphics[scale=0.2]{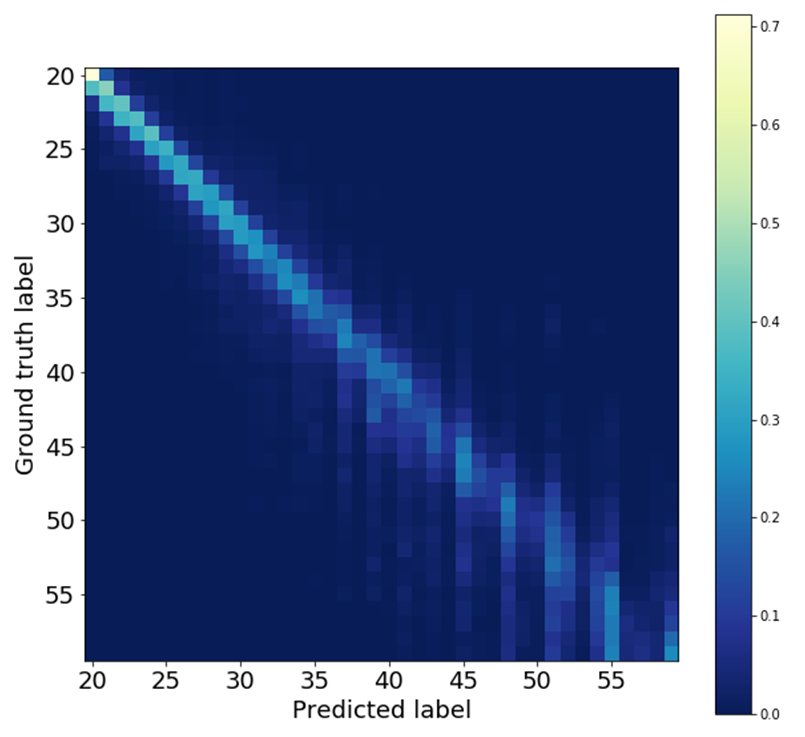}
\par\end{centering}

}

\caption{The plots of the normalized confusion matrices, where the predicted
labels are produced by (a) ordinal inference and (b) cross entropy.
Here we only show the depth labels between 20 and 60, as the samples
in this range is dominant and representative in KITTI dataset.\label{fig5}}

\end{figure}

To demonstrate the effectiveness of the ordinal loss, we compared
the depth estimation obtained by ordinal inference and that obtained
using cross entropy. The experiment is evaluated on KITTI dataset.
Normalized confusion matrices are plotted in Fig. \ref{fig5}. On
the plots of confusion matrices, the columns show the predicted depth
label, and the rows correspond to the true class. The diagonal elements
of the plots show what percentage of the pixels the trained network
correctly estimates their true classes. That is, it shows what percentage
of the true and predicted labels match. The off-diagonal elements
show where the depth estimation has made mistakes. From Fig. \ref{fig5},
it can be found that the ordinal-inference-based depth estimation
achieves higher estimation accuracy.

\begin{figure}
\begin{centering}
\includegraphics[scale=0.25]{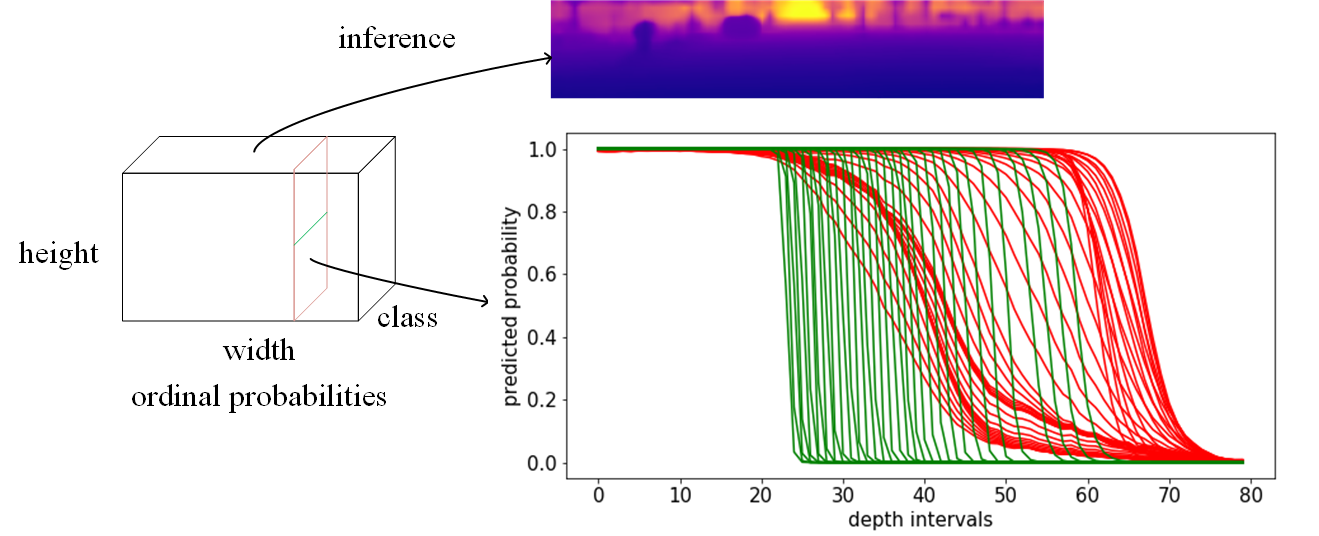}
\par\end{centering}
\caption{The typical probability distribution of the output of ordinal inference.
Each curve on the plot represents the predicted ordinal probability
set $\{\mathcal{P}_{i}^{0},\mathcal{P}_{i}^{1},\ldots,\mathcal{P}_{i}^{K}\}$
at position i.\label{fig6}}

\end{figure}

To illustrate the effectiveness of the proposed soft ordinal inference,
we give the output probabilities of the ACAN in Fig. \ref{fig6},
which is defined by Eq. \ref{eq:(10)}. In the inference phase, the
predicted depth map can be inferred from the output probability distribution.
Different position has different distributions of possible depth classes,
some of which are easy to estimate while others are not. For example,
in Fig. \ref{fig6}, the depth classes of the green curves in the
plot can be easily determined, while those of the red curves are hard
to distinguish clearly. One plausible explanation is that the model
for depth estimation has uncertainty to distinguish the depth classes
from its nearby intervals at these locations. Interestingly, the probability
distribution curves are roughly symmetric and centralized around the
right labels. Therefore, the area under the curve of probability distribution
is nearest to the ground truth depth label. Considering this statistical
analysis, we propose the soft ordinal inference to estimate the depth
values from the output probability effectively, which can not only
infer the correct depth class but also make up the decimal error of
quantization.

Furthermore, we then present both the qualitative and quantitative
experimental results. All these experiments are implemented with ResNet-50.
As illustrated in Table \ref{table3}, \textquoteleft CE(hard)\textquoteright{}
represents the model trained by cross entropy and applies hard-max
inference, while \textquoteleft CE(soft)\textquoteright{} applies
soft-weighted-sum inference proposed in \cite{Li2018Monocular}. \textquoteleft OR(hard)\textquoteright{}
represents the model trained by ordinal loss and applies hard-threshold
inference defined in Eq. \ref{eq:(12)}, while \textquoteleft OR(soft)\textquoteright{}
applies the proposed soft ordinal inference defined in Eq. \ref{eq:(13)}.
For fair comparison, the results of existing methods are also constructed
with ResNet-50.

It can be found that (1) No matter using the cross entropy loss or
ordinal loss, soft inference is always better than their hard counterparts.
(2) The RMSEs reduce clearly (NYU v2: 1.11\% for CE, 1.14\% for OR;
KITTI: 0.56\% for CE, 1.32\% for OR) while applying the soft inference.
(3) The proposed soft ordinal inference achieves the best result.

\begin{figure}
\begin{centering}
\includegraphics[scale=0.5]{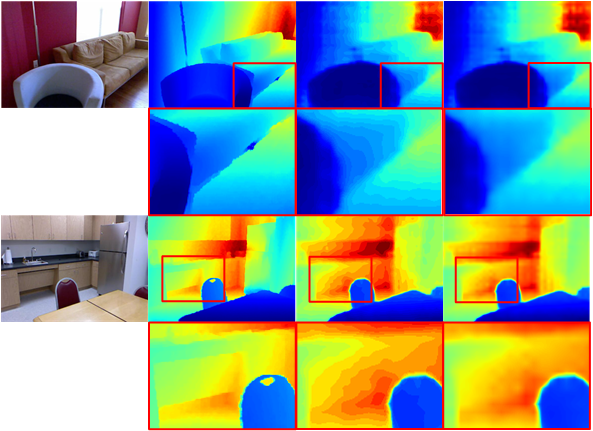}
\par\end{centering}
\caption{From left to right: input RGB image; ground truth; results of OR(hard);
results of OR(soft). The images in bottom rows show the details in
the red frames.\label{fig7}}

\end{figure}

\begin{table*}
\begin{centering}
\begin{tabular}{c|c|ccc|cccc}
\hline 
\multicolumn{2}{c|}{} & $\delta_{1}$ & $\delta_{2}$ & $\delta_{3}$ & RMSE & RMSE(log) & ARE & SRE\tabularnewline
\hline 
\hline 
\multirow{7}{*}{NYU} & Li \cite{Li2018Monocular} & 80.8\% & 95.7\% & 98.5\% & 0.601 & / & 0.147 & /\tabularnewline
 & Xu \cite{Xu2017Multi} & 81.1\% & 95.4\% & 98.7\% & 0.586 & / & \textbf{0.121} & /\tabularnewline
 & Laina \cite{laina2016deeper} & 81.1\% & 95.3\% & 98.8\% & 0.573 & 0.195 & 0.127 & /\tabularnewline
\cline{2-9} 
 & CE(hard) & 79.9\% & 95.6\% & 98.8\% & 0.536 & 0.188 & 0.151 & 0.118\tabularnewline
 & CE(soft) & 80.0\% & 95.7\% & 98.9\% & 0.530 & 0.187 & 0.150 & 0.115\tabularnewline
 & OR(hard) & 81.4\% & 96.0\% & 98.8\% & 0.524 & 0.183 & 0.147 & 0.114\tabularnewline
 & OR(soft) & \textbf{81.5\%} & \textbf{96.0\%} & \textbf{98.9\%} & \textbf{0.518} & \textbf{0.180} & 0.143 & \textbf{0.110}\tabularnewline
\hline 
\multirow{7}{*}{KITTI} & Godard \cite{Godard2017Unsupervised} & 86.1\% & 94.9\% & 97.6\% & 4.935 & 0.206 & 0.190 & 1.515\tabularnewline
 & Zhang \cite{Zhang2018Progressive} & 86.4\% & 96.6\% & 98.9\% & 4.082 & 0.164 & 0.139 & /\tabularnewline
 & Li \cite{Li2018Monocular} & 83.3\% & 95.6\% & 98.5\% & 5.325 & / & 0.128 & /\tabularnewline
\cline{2-9} 
 & CE(hard) & 86.9\% & 96.8\% & 99.1\% & 4.446 & 0.163 & 0.105 & 0.664\tabularnewline
 & CE(soft) & 87.0\% & 96.9\% & 99.2\% & 4.421 & 0.160 & 0.103 & 0.631\tabularnewline
 & OR(hard) & 91.5\% & 98.2\% & 99.5\% & 3.686 & 0.132 & 0.086 & 0.461\tabularnewline
 & OR(soft) & \textbf{91.5\%} & \textbf{98.3\%} & \textbf{99.5\%} & \textbf{3.637} & \textbf{0.130} & \textbf{0.085} & \textbf{0.445}\tabularnewline
\hline 
\multicolumn{1}{c}{} & \multicolumn{1}{c}{} & \multicolumn{3}{c|}{higher is better} & \multicolumn{4}{c}{lower is better}\tabularnewline
\hline 
\end{tabular}
\par\end{centering}
\caption{Comparisons of different training losses and inference strategies
On NYU and KITTI with ResNet-50 \label{table3}}

\end{table*}
The qualitative comparison of the hard inference and our soft ordinal
inference are illustrated in Fig.\ref{fig7}. As observed in Fig.
\ref{fig7}, the results of OR(hard) produce distorted predictions
on the transition area of depth while the results of OR(soft) are
smooth, continuous and similar to the ground truth depth map, which
well indicates that our soft ordinal inference can give the more realistic
depth map without introducing the stepped artifact. 

\subsubsection{Comparisons with the-state-of-the-arts}

1) NYU v2 depth: We compare our proposed ACAN with state-of-the-arts
on NYU v2 depth dataset. The results are shown in Table \ref{table4},
and the values in Table are copied from their respective papers directly.
In Table \ref{table4}, the \textquoteleft RX\textquoteright{} in
the brackets means the model is backboned on ResNet-X. 

As we observe, our model obtains the best performance among all of
the ResNet-50 based methods in all metrics and is even better than
some methods built on a more stronger backbone; our ResNet-101 based
model obtains competitive performance campared with some satae-of-the-arts.
Specifically, in terms of RMSE, our best model outperforms the previous
works in a large margin, as our quantization strategy and soft ordinal
inference greatly reduce the discretization error.

\begin{figure*}
\begin{centering}
\includegraphics[scale=0.6]{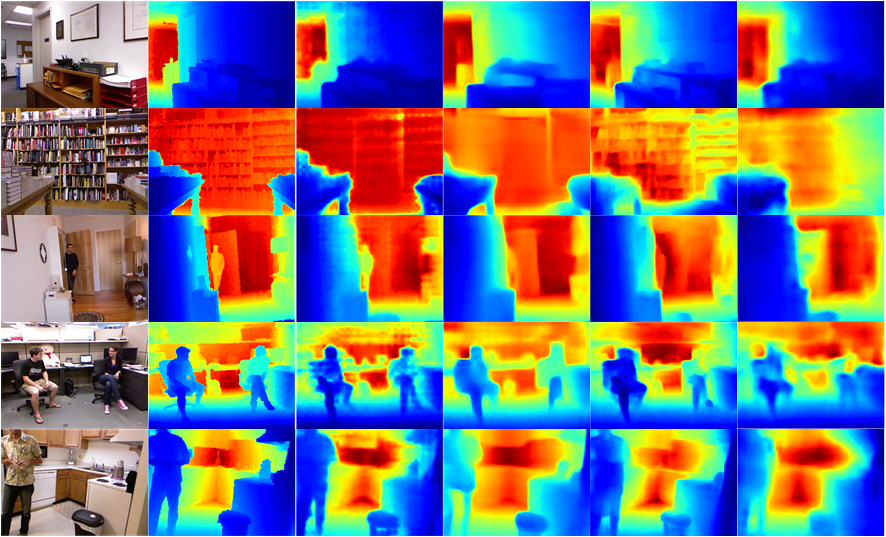}
\par\end{centering}
\caption{From left to right: input RGB images; ground truth; results of ours;
results of \cite{laina2016deeper}; results of \cite{Eigen2014Predicting};
results of \cite{Xu2017Multi}. \label{fig8}}

\end{figure*}

\begin{figure}
\begin{centering}
\includegraphics[scale=0.55]{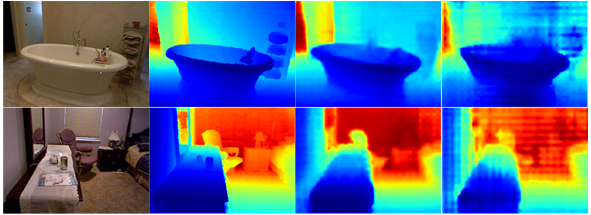}
\par\end{centering}
\caption{From left to right: input RGB images; ground truth; results of ours;
results of the ASPP-based method. \label{fig9}}

\end{figure}

\begin{table*}
\begin{centering}
\begin{tabular}{c|ccc|cccc}
\hline 
\multicolumn{1}{c|}{} & $\delta_{1}$ & $\delta_{2}$ & $\delta_{3}$ & RMSE & RMSE(log) & ARE & SRE\tabularnewline
\hline 
\hline 
Make3D \cite{Saxena2007Learning} & 44.7\% & 74.5\% & 89.7\% & 1.214 & / & / & /\tabularnewline
Ladicky \cite{Ladicky2014Pulling} & 54.2\% & 82.9\% & 94.1\% & / & / & / & /\tabularnewline
Liu \cite{Liu2015Deep} & 61.4\% & 88.3\% & 97.1\% & 0.824 & / & 0.230 & /\tabularnewline
Li \cite{Li2015Depth} & 62.1\% & 88.6\% & 96.8\% & 0.821 & / & 0.232 & /\tabularnewline
Roy \cite{Roy2016Monocular} & / & / & / & 0.744 & / & 0.187 & /\tabularnewline
Liu \cite{Liu2015Learning} & 65.0\% & 90.6\% & 97.6\% & 0.759 & / & 0.213 & /\tabularnewline
Eigen \cite{Eigen2014Depth} & 61.1\% & 88.7 & 97.1\% & 0.907 & 0.285 & 0.158 & 0.121\tabularnewline
Eigen \cite{Eigen2014Predicting} & 76.9\% & 95.0\% & 98.8\% & 0.641 & 0.214 & 0.158 & 0.121\tabularnewline
\hline 
Laina (R50) \cite{laina2016deeper} & 81.1\% & 95.3\% & 98.8\% & 0.573 & 0.195 & 0.127 & /\tabularnewline
Xu (R50) \cite{Xu2017Multi} & 81.1\% & 95.4\% & 98.7\% & 0.583 & / & \textbf{0.121} & /\tabularnewline
Li (R50) \cite{Li2018Monocular} & 80.8\% & 95.7 & 98.5\% & 0.601 & / & 0.147 & /\tabularnewline
\hline 
Li (R101) \cite{Li2018Monocular} & 82.0\% & 96.0\% & 98.9\% & 0.545 & / & 0.139 & /\tabularnewline
Yan (R101) \cite{yan2018monocular} & 81.3\% & 96.5\% & 99.3\% & 0.502 & / & 0.135 & /\tabularnewline
\hline 
Cao (R152) \cite{Cao2017Estimating} & 81.9\% & 96.5\% & 99.2\% & 0.540 & / & 0.141 & /\tabularnewline
Li (R152) \cite{Li2018Monocular} & \textbf{83.2\%} & 96.5\% & 98.9\% & 0.540 & 0.187 & 0.134 & \textbf{0.095}\tabularnewline
Moukari (R200) \cite{moukari2018deep} & 83.0\% & \textbf{96.6\%} & \textbf{99.3\%} & 0.569 & / & 0.133 & /\tabularnewline
\hline 
Our (R50) & 81.5\% & 96.0\% & 98.9\% & 0.518 & 0.180 & 0.144 & 0.110\tabularnewline
Our (R101) & 82.6\% & 96.4\% & 99.0\% & \textbf{0.496} & \textbf{0.174} & 0.138 & 0.101\tabularnewline
\hline 
\multicolumn{1}{c}{} & \multicolumn{3}{c|}{higher is better} & \multicolumn{4}{c}{lower is better}\tabularnewline
\hline 
\end{tabular}
\par\end{centering}
\caption{Comparisons between Our Proposed Method and The Different Previous
State-of-The-Arts on NYU v2 Depth Dataset \label{table4}}

\end{table*}
Qualitative results are illustrated in Fig. \ref{fig8} and Fig. \ref{fig9}.
As we can observe in Fig. \ref{fig8}, the results of \cite{laina2016deeper}
give the semantical predictions, as their method imposed the over-downsampling
to the feature maps and lack of the detail-reconstruction mechanism,
resulting in their predictions corrupt into the combination of simple
geometries. The results of \cite{Eigen2014Predicting} contain more
details but introduce certain distortion. For example, in the third
row of Fig. \ref{fig8}, the depth of the person is estimated inaccurate
in the result of \cite{Eigen2014Predicting}. The result of \cite{Xu2017Multi}
is blurry. In contrast, our results are detail-abundant and match
the ground truth well as a whole, as our attention model can extract
the global context features and is structure-aware.

Fig. \ref{fig9} shows that the results of the ASPP-based method will
introduce severe grid artifacts. The reason is that the kernels of
the ASPP-based method are predefined elaborately, which cannot adapt
to different objects in the image. However, the proposed ACAN method
can produce the piecewise smoothness depth map with more details visually.

2) KITTI: Table \ref{table5} shows the experimental results of the
proposed ACAN and the several state-of-the-art methods on KITTI dataset.

As we observe, our proposed ACAN (no matter using ResNet-50 or ResNet-101
as the encoder) achieves the excellent performance in all of the settings.
Moreover, ACAN (ResNet-50) outperforms the other methods even some
of their models are built on a stronger encoder. This can demonstrate
that our ACAN with soft ordinal inference is a more efficient method
for depth estimation.

\begin{figure*}
\begin{centering}
\includegraphics[scale=0.4]{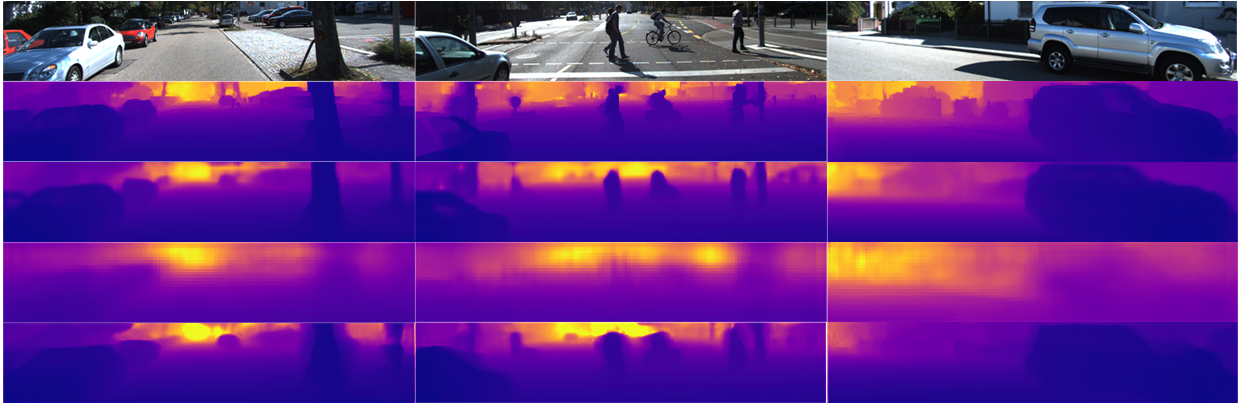}
\par\end{centering}
\caption{From up to bottom: input RGB images; ground truth; ours; results of
\cite{Eigen2014Depth}; results of \cite{Godard2017Unsupervised}.
\label{fig10}}
\end{figure*}

\begin{table*}
\begin{centering}
\begin{tabular}{c|ccc|cccc}
\hline 
\multicolumn{1}{c|}{} & $\delta_{1}$ & $\delta{}_{2}$ & $\delta_{3}$ & RMSE & RMSE(log) & ARE & SRE\tabularnewline
\hline 
\hline 
Liu \cite{Liu2015Learning} & 64.7\% & 88.2\% & 96.1\% & 6.986 & 0.289 & 0.217 & /\tabularnewline
Eigen \cite{Eigen2014Depth} & 69.2\% & 89.9\% & 96.7\% & 7.156 & 0.270 & 0.190 & 1.515\tabularnewline
Garg \cite{Garg2016Unsupervised} & 74.0\% & 90.4\% & 96.2\% & 5.104 & 0.273 & 0.169 & 1.080\tabularnewline
\hline 
Godard (R50) \cite{Godard2017Unsupervised} & 86.1\% & 94.9\% & 97.6\% & 4.935 & 0.206 & 0.114 & 0.898\tabularnewline
Zhang (R50) \cite{Zhang2018Progressive} & 86.4\% & 96.6\% & 98.9\% & 4.082 & 0.164 & 0.136 & /\tabularnewline
Li (R50) \cite{Li2018Monocular} & 83.3\% & 95.6\% & 98.5\% & 5.325 & / & 0.128 & /\tabularnewline
\hline 
Li (R101) \cite{Li2018Monocular} & 85.7\% & 96.5\% & 98.9\% & 4.528 & / & 0.106 & /\tabularnewline
Li (R152) \cite{Li2018Monocular} & 86.8\% & 96.7\% & 99.0\% & 4.513 & 0.164 & 0.104 & 0.697\tabularnewline
\hline 
Cao (R152) \cite{Cao2017Estimating} & 88.7\% & 96.3\% & 98.2\% & 4.712 & 0.198 & 0.115 & /\tabularnewline
\hline 
Our (R50) & 91.5\% & \textbf{98.3\%} & 99.5\% & 3.637 & 0.130 & 0.085 & 0.445\tabularnewline
Our (R101) & \textbf{91.9\%} & 98.2\% & \textbf{99.5\%} & \textbf{3.599} & \textbf{0.127} & \textbf{0.083} & \textbf{0.437}\tabularnewline
\hline 
\multicolumn{1}{c}{} & \multicolumn{3}{c|}{higher is better} & \multicolumn{4}{c}{lower is better}\tabularnewline
\hline 
\end{tabular}
\par\end{centering}
\caption{Comparisons between Our Proposed Method and Different State-of-The-Art
Method On KITTI Dataset \label{table5}}

\end{table*}
Qualitative results are illustrated in Fig. \ref{fig10}. As observed
in Fig. \ref{fig10}, the result of \cite{Eigen2014Depth} only give
the coarse and blurry predictions. The result of \cite{Godard2017Unsupervised}
are visually plausible, however, the depth maps of which are reconstructed
indirectly via learning the disparity of the given view under the
stereo constraint, which may introduce the noise. For example, the
predictions of the car and the tree are confused with the background
in the result of \cite{Godard2017Unsupervised}. In contrast, the
predictions of our method are visually satisfactory, where objects
of different scales can be recognized and our model can predict the
sharp boundaries as our attention model can capture the variable pixel-level
context adaptively.

\section{Conclusion}

In this paper, we propose a deep-CNN-based method, called the attention-based
context aggregation network (ACAN), for monocular depth estimation.
By utilizing the self-attention model, the proposed ACAN is able to
capture the long-range contextual information by learning the pixel-level
attention map adaptively, which is essential for the fine-grained
depth estimation. The image pooling module is also incorporated in
the ACAN, which can obtain the discriminative image-level context.
The aggregation of the pixel-level and image-level context is effective
to promote the performance of depth estimation. Soft ordinal inference
is also proposed in this paper, which takes full advantage of the
output ordinal probabilities to reduce the discretization error. The
experiments on NYU v2 dataset and KITTI dataset well demonstrate the
superiority of our model. In the future, we plan to investigate the
more effective variant of ACAN and extend our method to other dense
labeling tasks, such as semantic semantation and surface normal prediction.
Moreover, incorporating these tasks into the depth estimation is also
our interesting work.

\bibliographystyle{plain}
\bibliography{ref}

\end{document}